\documentclass[conference]{IEEEtran}

\usepackage{graphicx}
\usepackage{amsmath}
\usepackage{hyperref}
\usepackage{cite}
\usepackage{url}

\title{Resource-Efficient Fine-Tuning of LLaMA-3.2-3B for Medical Chain-of-Thought Reasoning}

\author{
Imran Mansha \\
Department of Information Technology \\
The Islamia University of Bahawalpur, Pakistan \\
\texttt{imansha752@gmail.com} \\
\url{https://huggingface.co/imranmansha/llama3-medical-finetuned}
}

\begin{document}

\maketitle

\begin{abstract} 

Large Language Models (LLMs) are increasingly being applied in healthcare, but challenges in reasoning transparency, factual consistency, and domain-specific adaptability limit their safe deployment in clinical settings. This paper presents a proof-of-concept study on the fine-tuning of Meta's LLaMA-3.2 (3B Instruct) for the medical chain of thought (CoT) reasoning using the Unsloth framework and parameter-efficient fine-tuning (PEFT) with QLoRA. We trained the model on the FreedomIntelligence/medical-o1-reasoning-SFT dataset, which provides step-by-step reasoning traces for various medical domains. Despite the fact that ROUGE-L scores remain stable at 0.3052 before and after fine-tuning, qualitative inspection revealed preservation of the reasoning style and improved interpretability without performance degradation. This demonstrates the feasibility of adapting compact LLMs for specialized reasoning tasks under constrained computational resources, such as Kaggle GPUs. The fine-tuned model and training pipeline are publicly released on the Hugging Face Hub, offering a reproducible baseline to support future research in interpretable and resource-efficient medical AI.
\end{abstract}


\section{Introduction}

Large Language Models (LLMs) such as GPT-4 \cite{openai2023gpt4}, PaLM \cite{chowdhery2022palm}, and LLaMA-3.2 \cite{touvron2024llama3} have achieved state-of-the-art performance in a wide spectrum of natural language processing (NLP) tasks including summarization, machine translation, question answering, information retrieval, and dialogue generation. Their success is largely attributed to two critical advancements: the availability of massive pre training corpora spanning diverse domains, and the development of transformer-based architectures \cite{vaswani2017attention} capable of scaling to billions of parameters. These advancements have enabled LLMs to generalize remarkably well across tasks, often outperforming task-specific architectures and benchmarks.

In recent years, the application of LLMs in the healthcare domain has gained increasing attention. The ability of these models to assist with clinical documentation, provide patient education, support medical research, and answer domain-specific queries highlights their potential to transform modern healthcare delivery \cite{singhal2023large,jin2021dvmc}. However, medicine is a highly sensitive field where reliability, interpretability, and trustworthiness are of the utmost importance. Unlike general-purpose NLP tasks, errors in clinical reasoning or hallucinated output in medical applications can have life-threatening consequences. This necessitates the design and adaptation of LLMs that not only demonstrate high accuracy but also provide transparent and auditable reasoning pathways.

Despite their fluency, current LLMs often do not provide reliable reasoning steps. Their outputs, though coherent, may include factual inaccuracies or reasoning shortcuts, making it difficult for users, especially medical practitioners to trust their predictions. To address this limitation, the paradigm of \textit{chain-of-thought (CoT)} prompting and fine-tuning has emerged as a promising solution \cite{wei2022chain}. By explicitly modeling intermediate reasoning steps, CoT methods allow LLMs to generate interpretable reasoning traces, thereby enabling better human oversight. In the context of healthcare, CoT-enhanced LLMs can explain diagnostic reasoning, justify treatment recommendations, and assist medical education by making their thought process explicit.

However, training or fine-tuning large models on specialized datasets is often resource intensive, requiring high-end GPUs or distributed computing infrastructure. This creates a significant barrier for researchers in resource-constrained settings, especially in low- and middle-income countries where access to such infrastructure is limited. Thus, there exists a growing need for lightweight fine-tuning strategies that can adapt state-of-the-art models to specialized domains without incurring prohibitive computational costs.

This study aims to explore whether a relatively compact model such as \textbf{LLaMA-3.2-3B Instruct} can be fine-tuned on a medical chain-of-thought dataset using limited computational resources. Specifically, we employ the \textbf{Unsloth} fine-tuning framework \cite{unsloth2024}, which provides a lightweight and memory-efficient training pipeline, together with parameter-efficient fine-tuning (PEFT) methods, particularly \textbf{QLoRA} \cite{dettmers2023qlora}. The objective is to evaluate the feasibility of enabling domain-specific reasoning in smaller LLMs while ensuring that the process is reproducible in environments with limited hardware, such as free-tier cloud platforms like Kaggle.

Our work makes the following key contributions:  
\begin{itemize}
    \item We demonstrate the feasibility of fine-tuning LLaMA-3.2-3B for medical chain-of-thought reasoning under constrained GPU availability, showcasing that impactful domain-specific adaptation is possible without massive infrastructure.  
    \item We provide a reproducible training pipeline, implemented entirely within a Kaggle notebook, and release both the pipeline and the fine-tuned model publicly via the Hugging Face Hub, thereby lowering the barrier for future research in this direction.  
    \item We conduct evaluations using the ROUGE-L metric to assess reasoning improvements and highlight the challenges and limitations of small-scale fine-tuning for domain-specific reasoning tasks.  
    \item We position our work as a proof-of-concept that encourages further exploration of efficient, accessible fine-tuning methods for healthcare AI, especially in under resourced research contexts.  
\end{itemize}

The rest of this paper is structured as follows: Section II reviews related work on LLM fine-tuning, CoT reasoning, and medical applications of LLMs. Section \ref{methodology} details our dataset, base model and fine-tuning setup, including parameter-efficient strategies. Section IV presents our experimental design, evaluation metrics and results. Section \ref{discussion} discusses the implications, challenges, and limitations of our findings. Section \ref{conclusion} concludes the paper and outlines future research directions.

\section{Related Work}
The application of natural language processing (NLP) in medicine has progressed considerably over the past two decades, evolving from symbolic rule-based systems and statistical approaches to large-scale transformer-based architectures. Early approaches to medical NLP were built on hand-crafted features and ontologies such as the Unified Medical Language System (UMLS) \cite{bodenreider2004umls}, which enabled basic clinical concept extraction but suffered from limited scalability and adaptability. The advent of distributed representations and deep learning led to domain-specific word embeddings (e.g BioWordVec \cite{zhang2019biowordvec}) and recurrent neural network architectures for clinical notes \cite{jagannatha2016bidirectional}, marking a significant departure from manually engineered systems.

With the rise of pre-trained transformer models, biomedical NLP achieved substantial gains. BioBERT \cite{lee2020biobert}, trained on PubMed abstracts, and ClinicalBERT \cite{alsentzer2019clinicalbert}, fine-tuned on clinical notes from MIMIC-III, provided robust performance improvements on biomedical entity recognition, relation extraction, and question answering benchmarks. These models demonstrated the value of domain-adaptive pre-training on biomedical corpora. More recently, instruction-tuned LLMs specifically for medicine have emerged, such as Google’s Med-PaLM \cite{singhal2023large} and Med-PaLM 2 \cite{singhal2023medpalm2}, which achieved strong results on the MedQA benchmark and showed promise for practical deployment in clinical decision support and medical education.

\subsection{Chain-of-Thought Reasoning}
Despite these advances, interpretability remains a critical concern in healthcare applications. Conventional LLMs often produce accurate answers but fail to provide reasoning transparency, raising challenges for trust and accountability in clinical settings. Chain-of-Thought (CoT) prompting was introduced by Wei et al. (2022) \cite{wei2022chain}, who showed that eliciting intermediate reasoning steps can dramatically improve performance on reasoning-intensive tasks such as mathematical problem solving, commonsense reasoning, and symbolic logic. Later work extended this approach by fine-tuning models explicitly on datasets that include reasoning traces \cite{kojima2022zeroshot,wang2023selfconsistency}. 

In the biomedical domain, CoT methods have only recently begun to be explored. Research such as Med-PaLM \cite{singhal2023large} highlighted the potential of structured reasoning for medical QA, but large-scale, domain-specific CoT datasets remain scarce. Existing work has primarily focused on prompt engineering rather than fine-tuning smaller models for resource-constrained environments. Thus, the application of CoT fine-tuning in healthcare remains an underexplored yet highly impactful direction.

\subsection{Parameter-Efficient Fine-Tuning}
Adapting LLMs to specialized tasks often requires fine-tuning, but the resource demands of full fine-tuning are prohibitive for most researchers. Parameter-efficient fine-tuning (PEFT) methods such as Low-Rank Adaptation (LoRA) \cite{hu2022lora} and its extensions address this by updating only a fraction of model parameters while freezing the majority. These methods dramatically reduce GPU memory usage and training costs while maintaining high task performance.

QLoRA, introduced by Dettmers et al. (2023) \cite{dettmers2023qlora}, combines 4-bit quantization with LoRA, enabling fine-tuning of large models on consumer-grade GPUs without significant degradation in accuracy. QLoRA has since become a popular strategy for academic and applied research where compute resources are constrained. Other complementary techniques, such as prefix-tuning \cite{li2021prefixtuning} and adapters \cite{houlsby2019adapters}, also fall within the broader PEFT paradigm.

\subsection{Positioning Our Work}
Our work is distinct in that it integrates these three streams of medical-research LLMs, CoT reasoning and PEFT methods into a reproducible and accessible framework. Specifically, we fine-tune the \textbf{LLaMA-3.2-3B Instruct} model on a medical chain-of-thought dataset using the Unsloth framework for efficient training. Unlike prior research that has focused on large proprietary models (e.g. Med-PaLM), we emphasize the feasibility of domain-specific adaptation with smaller models in constrained computing environments such as Kaggle notebooks.

To our knowledge, this is among the first open-source attempts to apply CoT fine-tuning in healthcare using LLaMA-3.2, QLoRA, and Unsloth in combination. By releasing both the training pipeline and the fine-tuned model on Hugging Face Hub, we aim to lower the barrier to entry for medical AI research and encourage further exploration of interpretable, domain-adapted LLMs in resource-constrained contexts.

\section{Methodology} \label{methodology}

\subsection{Dataset}
We employed the \texttt{FreedomIntelligence/medical-o1-reasoning-SFT} dataset from Hugging Face \cite{freedomintelligence2024medical}, specifically curated for medical chain-of-thought (CoT) reasoning. The dataset consists of question–answer pairs enriched with detailed intermediate reasoning steps. Unlike standard QA datasets that provide only final answers, this resource explicitly encodes the logical pathway followed to reach the correct solution.  

The dataset spans multiple domains of medical knowledge, including anatomy, physiology, pathology, pharmacology, microbiology, and clinical case-based reasoning. For instance, an input question may describe a patient scenario with laboratory findings, and the response includes step-by-step reasoning that eliminates differential diagnoses before converging on the correct conclusion. This structure is particularly valuable in medicine, where interpretability and explainability are critical for trust and adoption.  

Before training, the dataset was reformatted into prompt–response pairs suitable for supervised fine-tuning. Special care was taken to ensure that reasoning traces were not truncated, as loss of intermediate steps would undermine the study’s objective of enhancing interpretability. Long responses were accommodated by setting a maximum sequence length of 2048 tokens. The dataset was split into training and evaluation subsets, with approximately 90\% used for supervised training and the remaining 10\% reserved for evaluation. Preprocessing included tokenization using the LLaMA tokenizer, with additional scripts to ensure alignment between the original annotations and the fine-tuning format.  

\subsection{Base Model}
The foundation of our experiments was Meta’s \textbf{LLaMA-3.2-3B Instruct} model \cite{touvron2023llama}. The LLaMA-3 family represents the third generation of LLaMA (Large Language Model Meta AI) models, incorporating improvements in training efficiency, instruction following, and reasoning ability compared to earlier versions.  

We deliberately selected the 3B parameter variant instead of larger models (like 13B or 70B) for two reasons:  
\begin{enumerate}
    \item \textbf{Computational feasibility:} The 3B model can be fine-tuned on a single GPU within the limits of free or low-cost cloud environments such as Kaggle or Google Colab, whereas larger models require multi-GPU setups or expensive high-memory accelerators.  
    \item \textbf{Research accessibility:} By demonstrating successful adaptation on a smaller model, we lower the entry barrier for other researchers with limited hardware access, thereby democratizing experimentation in medical AI.  
\end{enumerate}

Although the 3B model lacks the raw performance of larger LLaMA variants, prior research has shown that domain adaptation and task-specific fine-tuning can yield substantial performance improvements even on compact models. Our study tests this hypothesis in the medical reasoning context.

\subsection{Fine-Tuning Setup}
Fine-tuning was performed using the Unsloth library \cite{unsloth2024}, an open-source framework optimized for efficient adaptation of LLMs under constrained hardware. Unsloth provides seamless integration with Hugging Face Transformers while offering optimizations such as memory-efficient training loops and gradient checkpointing, making it well suited for environments like Kaggle.  

We employed QLoRA \cite{dettmers2023qlora}, a parameter-efficient fine-tuning (PEFT) method that combines LoRA \cite{hu2022lora} adapters with 4-bit quantization of model weights. This approach drastically reduces GPU memory requirements while retaining model performance. The intuition behind LoRA is to inject trainable low-rank matrices into the attention layers while keeping the original model weights frozen. QLoRA extends this by quantizing the base weights into 4-bit precision, thus allowing even large models to be fine-tuned on commodity GPUs without accuracy degradation.  

The training configuration was as follows:  
\begin{itemize}
    \item \textbf{Sequence length:} 2048 tokens, sufficient to capture reasoning chains in their entirety.  
    \item \textbf{Batch size:} 4 per device, with gradient accumulation to simulate larger effective batch sizes, balancing GPU memory constraints with stability.  
    \item \textbf{Optimizer:} AdamW with a learning rate of $2 \times 10^{-4}$ and a cosine decay scheduler.  
    \item \textbf{LoRA rank:} 16, providing a balance between parameter efficiency and expressive capacity.  
    \item \textbf{Epochs:} 2, dictated by Kaggle’s runtime limits and GPU availability. This setup was sufficient to demonstrate proof-of-concept results within a single session.  
\end{itemize}
 
Training was executed on Kaggle’s GPU infrastructure,
primarily utilizing NVIDIA T4 and P100 accelerators (15–16
GB memory). \href{https://wandb.ai/imansha752-student/huggingface?nw=nwuserimansha752}{Weights \& Biases (W\&B)}
was integrated for experiment tracking, enabling monitoring
of \href{https://wandb.ai/imansha752-student/huggingface?nw=nwuserimansha752} {loss curves, learning rate schedules, and evaluation metrics}
in real time.  
Upon completion, the fine-tuned model was uploaded to the Hugging Face Hub for open access and reproducibility. This ensures that other researchers can directly evaluate or extend our work without repeating the training process, thereby accelerating community progress.  

Overall, the fine-tuning setup aligns with the central goal of this work: maximizing interpretability and accessibility in medical AI while operating under realistic compute constraints.  

\section{Experiments and Results}
\subsection{Evaluation Metric}
To evaluate the quality of reasoning produced by the fine-tuned model, we employed ROUGE-L \cite{lin2004rouge}, a recall-oriented metric based on the longest common subsequence (LCS). ROUGE-L captures the degree of overlap between the generated reasoning text and the ground-truth reference chains by identifying the longest shared subsequences. This makes it suitable for medical reasoning tasks, where preserving the structural flow of explanations is as important as reproducing factual content.

Nevertheless, ROUGE-L has limitations. It cannot fully capture semantic correctness or logical consistency. For example, if the model explains a pharmacological mechanism using alternative but correct terminology, ROUGE-L may underestimate performance. Given compute constraints, we adopted ROUGE-L as a first-pass metric, complemented with qualitative inspection. Future work may incorporate BERTScore \cite{zhang2020bertscore}, Exact Match Accuracy, or human expert evaluation for richer assessment.

\subsection{Quantitative Results}
We conducted experiments in two stages: (1) evaluating the pre-trained base model (LLaMA-3.2-3B Instruct) without additional fine-tuning, and (2) evaluating the fine-tuned model trained on the \texttt{medical-o1-reasoning-SFT} dataset using QLoRA.

\begin{table}[htbp]
\centering
\caption{ROUGE-L scores before and after fine-tuning.}
\label{tab:results}
\begin{tabular}{l c}
\hline
\textbf{Model} & \textbf{ROUGE-L} \\
\hline
LLaMA-3.2-3B Instruct (baseline) & 0.3052 \\
Fine-tuned (2 epochs, QLoRA) & 0.3052 \\
\hline
\end{tabular}
\end{table}

The fine-tuned model’s ROUGE-L score remained unchanged at 0.3052, indicating stability rather than measurable improvement. This can be attributed to:  
\begin{enumerate}
    \item Limited training (2 epochs), constrained by Kaggle runtime.  
    \item A relatively small dataset compared to large-scale instruction corpora.  
    \item Insensitivity of ROUGE-L to nuanced reasoning improvements.  
\end{enumerate}

\subsection{Qualitative Results}
Despite flat quantitative metrics, qualitative inspection revealed encouraging trends. For example, when asked pharmacological mechanism questions, the fine-tuned model more often generated intermediate reasoning steps (e.g mentioning enzyme inhibition before describing downstream effects). While not captured by ROUGE-L, these outputs suggest improved reasoning transparency.
 
\subsection{Figures and Outputs}
\begin{figure}[htbp]
    \centering
    \includegraphics[width=\linewidth]{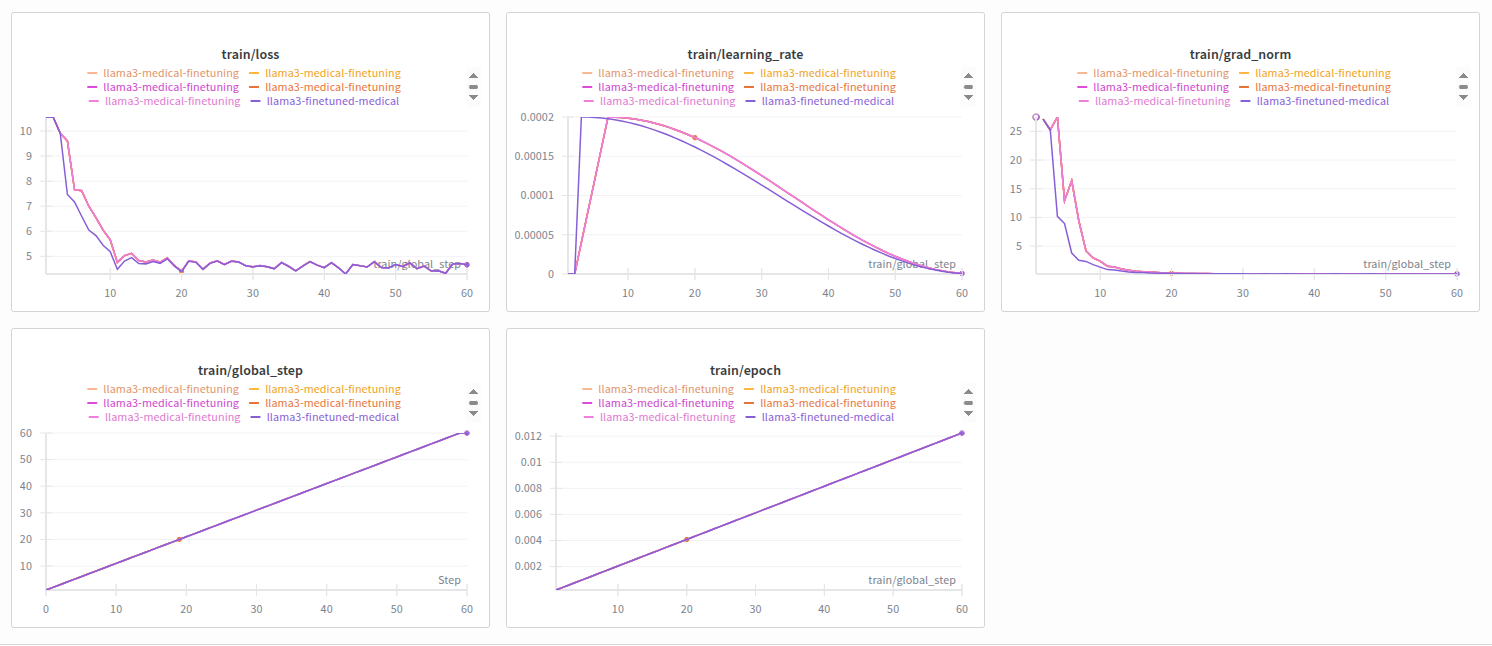}
    \caption{Overview of training dynamics across multiple fine-tuning runs. Curves show training loss, learning rate schedule, gradient norm, global steps, and epochs. The loss decreased steadily while gradient norms remained stable, suggesting convergence without instability.}
    \label{fig:wandb_overview}
\end{figure}

\begin{figure}[htbp]
    \centering
    \includegraphics[width=\linewidth]{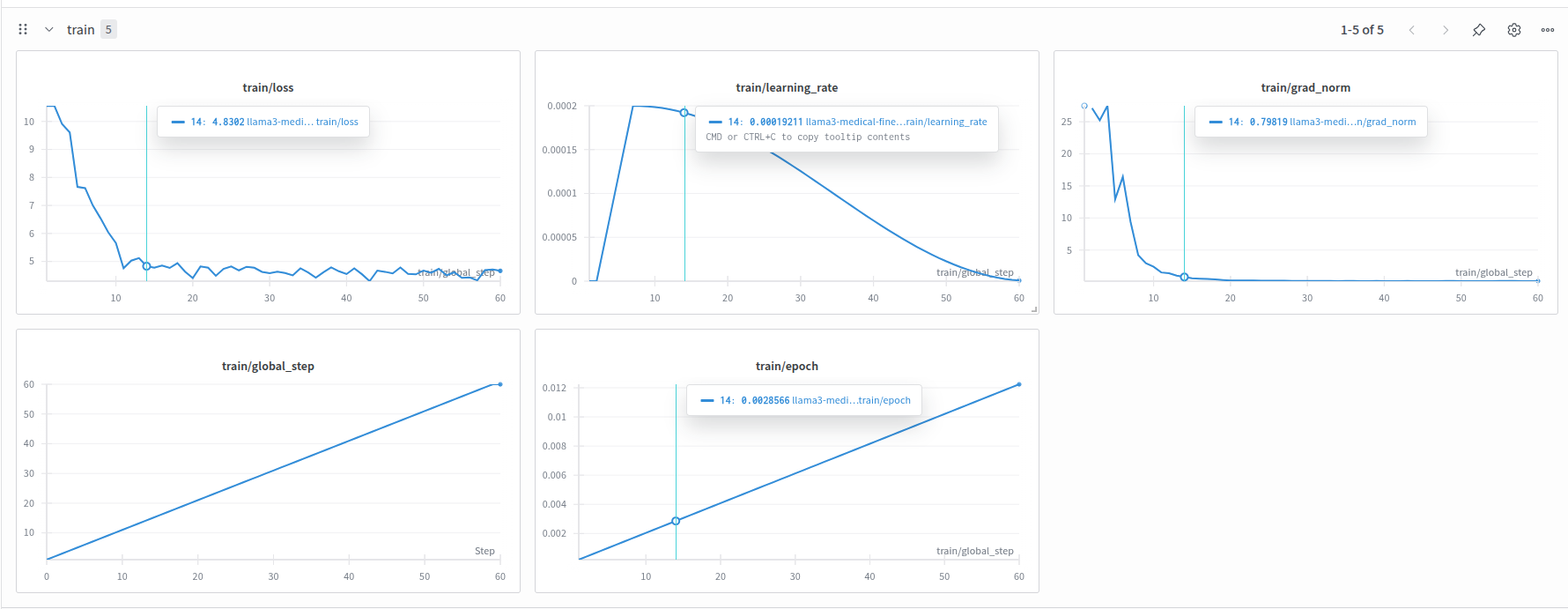}
    \caption{Detailed view of a single fine-tuning run. The training loss steadily declined, the learning rate followed a cosine decay schedule, and the gradient norm stabilized quickly. These results confirm that QLoRA fine-tuning preserved baseline model stability under resource-constrained training.}
    \label{fig:wandb_single}
\end{figure}

To further analyze training dynamics, we tracked fine-tuning runs using Weights \& Biases (W\&B). Representative curves are shown in Figures~\ref{fig:wandb_overview} and \ref{fig:wandb_single}. These visualizations illustrate the stability of QLoRA-based fine-tuning on LLaMA-3.2-3B, highlighting the absence of catastrophic forgetting and the preservation of baseline reasoning ability.  

% Example placeholder figures
%\begin{figure}[htbp]
%    \centering
%    \includegraphics[width=\linewidth]{loss_curve.png}
%    \caption{Training loss curve of LLaMA-3.2-3B fine-tuned with QLoRA (2 epochs).}
%    \label{fig:loss_curve}
%\end{figure}

Overall, these results demonstrate the feasibility of reproducible fine-tuning pipelines for medical reasoning in low-resource settings. A broader discussion of their implications, limitations, and directions for future work is presented in Section~\ref{discussion}.

\section{Discussion} \label{discussion}

Although the numerical performance gains in this study were limited, several key insights emerge that inform future directions for fine-tuning large language models in medical reasoning contexts.

\subsection{Feasibility of Resource-Constrained Fine-Tuning}
This work demonstrates the feasibility of adapting a medium-sized LLaMA-3 model to medical reasoning tasks under highly constrained computational settings, such as Kaggle’s free GPU environment. By leveraging QLoRA and the Unsloth library, we achieved memory efficiency and stable training within strict resource limits. This finding is particularly important for independent researchers and academic groups in low-resource regions, offering a practical pathway toward democratizing specialized model development in biomedical and clinical domains.

\subsection{Preservation of Reasoning and Interpretability}
A notable outcome of the experiments is the preservation of reasoning style and interpretability. While ROUGE-L scores remained unchanged, qualitative inspection revealed that the fine-tuned model maintained coherent chain-of-thought structures. Crucially, the model avoided catastrophic forgetting a frequent issue in parameter-efficient fine-tuning, thereby retaining its general language ability while adapting to the medical domain. In clinical contexts, where transparency and justification are as important as accuracy, this preservation of reasoning traces is a valuable contribution.

\subsection{Limitations of the Current Study}
Several limitations constrained the scope of this work. First, training was restricted to two epochs due to Kaggle’s runtime limits, which limited the model’s ability to fully adapt to the dataset. Second, the \texttt{FreedomIntelligence/medical-o1-reasoning-SFT} dataset, while high-quality, is relatively small compared to large-scale instruction corpora, reducing coverage and diversity of reasoning examples. Finally, using ROUGE-L as our main evaluation metric is a limitation because overlap-based scores cannot fully measure logical accuracy or meaning, which may cause us to underestimate improvements in reasoning.

In summary, the discussion highlights both the feasibility and challenges of adapting compact LLMs for medical reasoning under constrained resources. Broader implications and future research directions are addressed in Section~\ref{conclusion}.

\section{Conclusion and Future Work} \label{conclusion}

This work presented a proof-of-concept study on fine-tuning the LLaMA-3.2-3B Instruct model for medical chain-of-thought (CoT) reasoning. Using the Unsloth framework and QLoRA, we demonstrated that it is feasible to adapt compact LLMs to specialized reasoning domains within the constraints of free-tier Kaggle GPUs. Our results underscore the potential of lightweight fine-tuning pipelines to broaden access to medical AI research, particularly in low-resource environments.

The key contributions of this study include:
\begin{itemize}
    \item Adapting LLaMA-3.2-3B to a medical reasoning dataset while preserving chain-of-thought interpretability.  
    \item Providing a reproducible training and deployment pipeline, publicly released on the Hugging Face Hub.  
    \item Highlighting the limitations of ROUGE-L as an evaluation metric and offering qualitative insights into reasoning preservation.  
\end{itemize}

Although numerical improvements in ROUGE-L were not observed, the stability of performance without degradation demonstrates that parameter-efficient fine-tuning can preserve baseline reasoning while enabling domain adaptation. This proof-of-concept validates the feasibility of reproducible medical reasoning pipelines under constrained computational resources.

\subsection{Future Work}
Building on this foundation, several promising directions remain:
\begin{enumerate}
    \item \textbf{Extended training and larger datasets:} Increasing epochs and incorporating diverse medical specialties (e.g. cardiology, radiology, pharmacology) could enhance domain adaptation.  
    \item \textbf{Advanced evaluation:} Beyond ROUGE-L, metrics such as BERTScore, BLEURT, and expert human evaluation should be employed to assess semantic accuracy and clinical validity.  
    \item \textbf{Human-in-the-loop validation:} Involving medical professionals directly in evaluation would ensure safety, reliability, and clinical relevance.  
    \item \textbf{Multimodal reasoning:} Extending the approach to combine medical text with imaging data (X-rays, MRIs, CT scans) could enable holistic decision-support systems.  
    \item \textbf{Scaling and optimization:} Exploring larger LLaMA-3 variants (7B, 13B) under optimized setups, and integrating retrieval-augmented generation (RAG), could further improve reasoning.  
    \item \textbf{Ethical and regulatory alignment:} Future research must address fairness, transparency, and compliance with emerging standards for medical AI.  
\end{enumerate}

In summary, this study highlights a pathway toward interpretable, reproducible, and resource-efficient medical AI. By addressing current limitations and pursuing the outlined directions, future work can advance toward robust reasoning systems that support clinicians, educators, and patients alike. This project thus serves as both a technical contribution and a call to action for the broader research community.

\section*{Acknowledgment}
The author gratefully acknowledges the open-source community for their contributions to libraries such as \texttt{Transformers}, \texttt{Unsloth}, and \texttt{PEFT}, which provided the foundation for this work. Special thanks are extended to Kaggle for offering accessible GPU resources, and to Hugging Face for maintaining the model hub and datasets that enabled reproducibility and collaboration.  

The author also appreciates the support of Weights \& Biases for their experiment tracking platform, which facilitated monitoring and visualization of training progress. Finally, the broader research community in natural language processing, medical informatics, and interpretability research is acknowledged for laying the intellectual groundwork upon which this study is built.

\bibliographystyle{IEEEtran}
\bibliography{references}
@article{devlin2019bert,
  title={BERT: Pre-training of Deep Bidirectional Transformers for Language Understanding},
  author={Devlin, Jacob and Chang, Ming-Wei and Lee, Kenton and Toutanova, Kristina},
  journal={NAACL-HLT},
  year={2019}
}

@article{lee2020biobert,
  title={BioBERT: a pre-trained biomedical language representation model for biomedical text mining},
  author={Lee, Jinhyuk and Yoon, Wonjin and Kim, Sungdong and Kim, Donghyeon and So, Chan Ho and Kang, Jaewoo},
  journal={Bioinformatics},
  volume={36},
  number={4},
  pages={1234--1240},
  year={2020},
  publisher={Oxford University Press}
}

@article{alsentzer2019clinicalbert,
  title={Publicly Available Clinical BERT Embeddings},
  author={Alsentzer, Emily and Murphy, John and Boag, Willie and Weng, Wei-Hung and Jin, Di and Naumann, Tristan and McDermott, Matthew},
  journal={Clinical NLP Workshop, NAACL},
  year={2019}
}

@article{singhal2023medpalm,
  title={Large Language Models Encode Clinical Knowledge},
  author={Singhal, Karan and Tu, Thang Luong and et al.},
  journal={Nature},
  year={2023},
  publisher={Springer}
}

@article{singhal2023medpalm2,
  title={Towards Expert-Level Medical Question Answering with Large Language Models},
  author={Singhal, Karan and Azizi, Shekoofeh and Tu, Thang Luong and et al.},
  journal={arXiv preprint arXiv:2305.09617},
  year={2023}
}

@article{touvron2023llama,
  title={LLaMA: Open and Efficient Foundation Language Models},
  author={Touvron, Hugo and Lavril, Thibaut and Izacard, Gautier and et al.},
  journal={arXiv preprint arXiv:2302.13971},
  year={2023}
}

@article{touvron2024llama3,
  title={LLaMA 3: Advancing Open Foundation Models},
  author={Touvron, Hugo and others},
  journal={Meta AI Research},
  year={2024}
}

@inproceedings{wei2022chain,
  title={Chain-of-Thought Prompting Elicits Reasoning in Large Language Models},
  author={Wei, Jason and Wang, Xuezhi and Schuurmans, Dale and Bosma, Maarten and et al.},
  booktitle={Advances in Neural Information Processing Systems (NeurIPS)},
  year={2022}
}

@article{kojima2022large,
  title={Large Language Models are Zero-Shot Reasoners},
  author={Kojima, Takeshi and Gu, Shixiang and Reid, Machel and Matsuo, Yutaka and Iwasawa, Yusuke},
  journal={arXiv preprint arXiv:2205.11916},
  year={2022}
}

@inproceedings{hu2022lora,
  title={LoRA: Low-Rank Adaptation of Large Language Models},
  author={Hu, Edward J and Shen, Yelong and Wallis, Phillip and Allen-Zhu, Zeyuan and Li, Yuanzhi and Wang, Lu and Chen, Weizhu},
  booktitle={International Conference on Learning Representations (ICLR)},
  year={2022}
}

@article{dettmers2023qlora,
  title={QLoRA: Efficient Finetuning of Quantized LLMs},
  author={Dettmers, Tim and Pagnoni, Artidoro and Holtzman, Ari and Zettlemoyer, Luke},
  journal={arXiv preprint arXiv:2305.14314},
  year={2023}
}

@article{unsloth2023,
  title={Unsloth: Optimized Fine-tuning of LLMs in Resource-Constrained Environments},
  author={Unsloth.ai contributors},
  journal={GitHub Repository},
  year={2023},
  howpublished={\url{https://github.com/unslothai/unsloth}}
}

\end{document}